\documentclass[3p,times]{elsarticle}

\usepackage{hyperref}
\hypersetup{colorlinks=true,linkcolor=blue,urlcolor=blue}

\usepackage[utf8]{inputenc} 
\usepackage[T1]{fontenc} 
\usepackage{hyperref}       
\usepackage{url}            
\usepackage{booktabs}      
\usepackage{amsfonts}      
\usepackage{nicefrac}      
\usepackage{microtype}    
\usepackage{algorithmic}
\usepackage[ruled,vlined]{algorithm2e}
\usepackage{amsmath}
\usepackage{graphicx}
\usepackage{multicol}
\usepackage[font=footnotesize, labelfont=bf]{caption}
\DeclareMathOperator{\Tr}{Tr}
\usepackage{amsthm}
\newtheorem{theorem}{Theorem}
\newtheorem{corollary}{Corollary}[theorem]
\newtheorem{lemma}[theorem]{Lemma}
\newtheorem*{remark}{Remark}
\newtheorem{assumption}{Assumption}

\usepackage{multirow}
\usepackage{bbm}

\usepackage{pifont}

\usepackage{hyperref}
\hypersetup{colorlinks=true,linkcolor=blue,urlcolor=blue}

\usepackage[]{algorithm2e}
\SetKwInput{KwInput}{Input \hspace*{0.3em}}
\SetKwInput{KwOutput}{Output}

\usepackage{tikz}

\usepackage{amssymb}
\usepackage{pgfplots}
\usepackage{pgfmath}
\usepgfplotslibrary{patchplots}
\usetikzlibrary{patterns, positioning, arrows}
\usepackage[shortlabels]{enumitem}

\pgfmathsetmacro\sprayRadius{.75pt}
\pgfmathsetmacro\sprayPeriod{.8cm}

\usepackage{chngcntr}
\counterwithin{paragraph}{section}

\usepackage{amssymb}

\usepackage[figuresright]{rotating}

\begin{document}

\begin{frontmatter}




\title{Deep Optimal Transport for Domain Adaptation on SPD Manifolds}


\author{Ce Ju\corref{cor1}}
\ead{juce0001@ntu.edu.sg}
\ead{ce.ju@inria.fr}
\address{College of Computing and Data Science, Nanyang Technological University, 50 Nanyang Avenue, Singapore.} 
\address{Université Paris-Saclay, Inria, CEA, Palaiseau, France.}

\author{Cuntai Guan\corref{cor2}}
\ead{ctguan@ntu.edu.sg}
\address{College of Computing and Data Science, Nanyang Technological University, 50 Nanyang Avenue, Singapore.} 

\cortext[cor1]{Corresponding Author}
\cortext[cor2]{Principal Corresponding Author}

\begin{abstract}
Recent progress in geometric deep learning has drawn increasing attention from the machine learning community toward domain adaptation on symmetric positive definite (SPD) manifolds—especially for neuroimaging data that often suffer from distribution shifts across sessions. These data, typically represented as covariance matrices of brain signals, inherently lie on SPD manifolds due to their symmetry and positive definiteness. However, conventional domain adaptation methods often overlook this geometric structure when applied directly to covariance matrices, which can result in suboptimal performance. To address this issue, we introduce a new geometric deep learning framework that combines optimal transport theory with the geometry of SPD manifolds. Our approach aligns data distributions while respecting the manifold structure, effectively reducing both marginal and conditional discrepancies. We validate our method on three cross-session brain-computer interface datasets—KU, BNCI2014001, and BNCI2015001—where it consistently outperforms baseline approaches while maintaining the intrinsic geometry of the data. We also provide quantitative results and visualizations to better illustrate the behavior of the learned embeddings. 
\end{abstract}

\end{frontmatter}

\section{Introduction}

Cross-domain problems arise when the distribution of source data diverges from that of target data, a challenge widely studied as domain adaptation in machine learning~\cite{ben2007analysis,pan2009survey,wang2018deep}. For instance, even if signals in an online testing set and a calibration set originate from the same generative process, their distributions often differ significantly.
Formally, a domain $\mathcal{D}$ is defined by a feature space $X$ and its marginal probability distribution $\mathbb{P}(X)$, denoted as $\mathcal{D} = \{X, \mathbb{P}(X)\}$. A task $\mathcal{T}$ consists of a label space $Y$ and the conditional distribution $\mathbb{Q}(Y|X)$, expressed as $\mathcal{T} = \{Y, \mathbb{Q}(Y|X)\}$.
Domain adaptation aims to transfer knowledge from a source domain $\mathcal{D}_S$ and task $\mathcal{T}_S$ to improve model performance on a target domain $\mathcal{D}_T$ and task $\mathcal{T}_T$, where $\mathcal{D}_S \neq \mathcal{D}_T$ and $\mathcal{T}_S \neq \mathcal{T}_T$, specifically as follows:
\begin{itemize}
\item $\mathcal{D}_S \neq \mathcal{D}_T$ implies either $X_S \neq X_T$ (e.g., differing feature dimensions) or $\mathbb{P}_S(X_S) \neq \mathbb{P}_T(X_T)$ (e.g., covariate shift).
\item  $\mathcal{T}_S \neq \mathcal{T}_T$ indicates either $Y_S \neq Y_T$ (e.g., label space mismatch) or $\mathbb{Q}_S(Y_S|X_S) \neq \mathbb{Q}_T(Y_T|X_T)$ (e.g., label shift).
\end{itemize}

Domain adaptation is typically studied under two settings: unsupervised, where no target labels are available, and semi-supervised, where a small number of labeled target samples are accessible. EEG signals provide a classic example of domain shift: changes between calibration (training) and feedback (testing) sessions often degrade model performance. This has motivated adaptation methods such as covariate shift correction~\cite{sugiyama2007covariate}, which assume only that the marginal distribution $\mathbb{P}(X)$ changes, while $\mathbb{Q}(Y|X)$ remains stable. Optimal Transport (OT) provides a principled framework for domain adaptation by aligning distributions through the minimization of the transport cost, typically measured by the Wasserstein distance. In OT-based domain adaptation (OT-DA), the goal is to learn a push-forward map $\varphi:X_S \longmapsto X_T$ such that the pushforward distribution $\varphi_{\sharp}\mathbb{P}_S$ matches the target distribution $\mathbb{P}_T$. This is commonly achieved through regularized OT, allowing models trained in the source domain to generalize to the target domain while respecting the geometry of the data~\cite{courty2016optimal}.

In Electroencephalogram (EEG)-based motor imagery classification, signals are commonly represented as spatial covariance matrices averaged over trials~\cite{koles1990spatial,muller1999designing}. Equipped with a Riemannian metric such as the Log-Euclidean metric, these covariance matrices lie on the manifold of SPD matrices, forming a smooth Riemannian manifold with a well-defined geometric structure~\cite{arsigny2005fast,arsigny2006log}. This structure provides a principled foundation for developing learning algorithms on manifolds~\cite{barachant2011multiclass}. When adapted to this geometry, OT-DA becomes particularly effective. For instance, Yair et al.~\cite{yair2019domain} pioneered OT-DA on SPD manifolds by employing a squared Euclidean cost $c(x, \bar{x})=d_{\ell_2}(x, \bar{x})^2$, and derived a closed-form solution based on Brenier’s polar factorization. However, their method is limited by both theoretical soundness and practical performance:
\begin{enumerate}[i.]
\item Euclidean cost fails to reflect the intrinsic Riemannian geometry of SPD manifolds.
\item The method presumes a stable $\mathbb{Q}(Y|X)$, overlooking label shift. Additionally, label scarcity in test phases and frequency-band variability pose further modeling challenges.
\end{enumerate}
These limitations and challenges motivate the need for a more principled OT-DA framework that respects SPD geometry and accounts for both marginal and conditional distribution mismatches.

In this study, we focus on a domain adaptation scenario where the feature and label spaces are consistent ($X_S = X_T$, $Y_S = Y_T$), but both the marginal and conditional distributions differ. This setting calls for joint distribution adaptation, which simultaneously aligns $\mathbb{P}(X)$ and $\mathbb{Q}(Y|X)$, even when target labels are unavailable~\cite{long2013transfer}. To model $X_S = X_T$ on the SPD manifold $\mathcal{S}_{++}$, we use the Log-Euclidean metric ($g^{LEM}$)~\cite{arsigny2005fast}, which maps SPD matrices to a Euclidean tangent space at the identity matrix via matrix logarithms. This enables standard operations in the tangent space while preserving manifold structure. Notably, parallel transport under $g^{LEM}$ becomes identity, simplifying source-target alignment within a shared tangent space.

We propose Deep Optimal Transport (DOT), a geometric deep learning framework for joint domain adaptation on SPD manifolds that leverages a deep encoder to implicitly learn a nonlinear mapping on SPD manifolds as a geometric transformation. 
Specifically, the encoder-based paradigm projects source and target domains into a shared SPD subspace $Z \subset \mathcal{S}_{++}$ via $\varphi: X_S, X_T \longmapsto Z$, aligning both marginal and conditional distributions through a composite optimal transport loss.
The alignment is driven by minimizing the squared Log-Euclidean distance as the OT cost function, i.e.,  $c(x, \bar{x})=d_{g^{LEM}}^2(x, \bar{x})$, under the principle that domain discrepancies can be quantified and bridged through OT-based distribution matching. 
This enables $\mathbb{P}_S (\varphi(X_S)) \approx \mathbb{P}_T(\varphi(X_T))$ and $\mathbb{Q}_S(Y_S|\varphi(X_S)) \approx \mathbb{Q}_T(Y_T|\varphi(X_T))$ without assuming Gaussianity or linearity. 
We evaluate DOT on cross-session motor imagery classification, a challenging task where neuronal activity shifts over time, causing significant EEG signal variability across sessions. Our experiments span unsupervised and semi-supervised scenarios: the model is calibrated on one session and tested on another, demonstrating robustness to nonstationary brain dynamics. Unlike linear Wasserstein barycenter methods, DOT’s deep encoder captures nonlinear SPD embeddings, and the composite OT loss jointly addresses both domain-level and class-level distribution shifts.

The remainder of this paper is organized as follows: Section~\ref{preliminary} introduces basic concepts, including Log-Euclidean geometry, discrete OT, joint distribution alignment, and SPD matrix-based classifiers for motor imagery tasks. Section~\ref{methodology} details the proposed OT-DA framework on SPD manifolds and presents the architecture of DOT. Section~\ref{experiments} applies DOT to EEG-based motor imagery classification across diverse datasets. Section~\ref{sec:main_results} analyzes results from synthetic and real-world experiments. A comparative summary of existing frameworks is provided in Table~\ref{tab:comarison}, highlighting key methodological distinctions.

\begin{table*}[!t]
\caption{A comparison of various OT-DA Frameworks. $X_S$ and $X_T$ are feature spaces of the source and target domains, respectively. $\mathbb{P}$ and $\mathbb{Q}$ represent the marginal probability distribution and conditional probability distribution, respectively. 
~\label{tab:comarison}
}
\centering 
\begin{tabular}{l  l  l    l  l}
\toprule
Framework & Cost function & Space & Transformation &Domain Adaptation Scenario \\
\midrule
OT-DA\cite{courty2016optimal} & $c(x, \bar{x}) = d_{\ell_2}^2 (x, \bar{x}) $  & Euclidean & Affine Function & $X_S \neq X_T$.\\
JDOT-DA\cite{courty2017joint,damodaran2018deepjdot} & $c(x, \bar{x}) = d_{\ell_2}^2 (x, \bar{x}) $ & Euclidean & Affine Function & i). $X_S \neq X_T$;\\
&&&&$ii). \mathbb{Q}_S(Y_S|X_S) \neq \mathbb{Q}_T(Y_T|X_T)$.\\
OT-DA on $\mathcal{S}_{++}$\cite{yair2019domain}& $c(x,\bar{x}) = d_{\ell_2}^2 (x, \bar{x}) $ & ($\mathcal{S}_{++}$, $g^{AIRM}$) & BiMap Function & $X_S \neq X_T$.\\
DOT (Proposed)  & $c(x, \bar{x}) = d_{g^{LEM}}^2(x, \bar{x})$ & ($\mathcal{S}_{++}$, $g^{LEM}$) & Neural Network & i). $X_S = X_T$;\\
 &&&& ii). $\mathbb{P}_S(X_S) \neq \mathbb{P}_T(X_T);$\\
 && && iii). $\mathbb{Q}_S(Y_S|X_S) \neq \mathbb{Q}_T(Y_T|X_T)$.\\
\bottomrule
\end{tabular}
\end{table*}

\section{PRELIMINARY}~\label{preliminary}

\subsection*{Notations and Conventions}
A matrix $S \in \mathbb{R}^{n \times n}$ is symmetric and positive-definite if it is symmetric and all its eigenvalues are positive, i.e., $S = S^T$, and $x^TSx > 0$, for all $x \in \mathbb{R}^n \setminus \{{\mathbf{0}}\}$.
The \emph{Frobenius} inner product and \emph{Frobenius} norm on $m \times n$ matrices $A$ and $B$ are defined as $\langle \, A, B \, \rangle_{\mathcal{F}} := \Tr{(A^\top \cdot B)}$ and $||\,A\,||_{\mathcal{F}}^2 :=  \langle \, A, A \, \rangle_{\mathcal{F}}$ respectively.

\subsection{Riemannian Metric for SPD Manifolds}~\label{Appen:metrics}
The Log-Euclidean metric ($g^{LEM}$), a widely adopted Riemannian metric on SPD manifolds, induces a vector space structure on the space of SPD matrices, enabling Euclidean-like operations in the logarithmic domain while preserving geometric properties~\cite{arsigny2005fast,arsigny2006log}. 
Formally, a commutative Lie group structure, known as the tensor Lie group, is endowed on $\mathcal{S}_{++}$ using the logarithmic multiplication $\odot$ defined as follows, 
\[
P_1 \odot P_2:=\exp\big( \log P_1 + \log P_2\big),
\]
where $P_1, P_2 \in \mathcal{S}_{++}$, and $\exp$ and $\log$ are matrix exponential and logarithm respectively. 
The bi-invariant metric exists on the tensor Lie group, which yields the scalar product as follows,
\[
g^{LEM}_P (v, w):= \langle \, D_P \log v, D_P \log w \, \rangle_{\mathcal{F}},
\] 
where tangent vectors $v, w$ at matrix $P$, and $D_P \log$ represent the differential $D \log$ at $P$. 
We denote an SPD manifold equipped with the Log-Euclidean metric as ($\mathcal{S}_{++}$, $g^{LEM}$).
The sectional curvature of ($\mathcal{S}_{++}$, $g^{LEM}$) is flat, and thus parallel transport on it simplifies to an identity. 
Riemannian distance $d_{g^{LEM}} (P_1, P_2) = ||\log{P_1} - \log{P_2}||_{\mathcal{F}}$ between $P_1$ and $P_2$ on this SPD manifold. 
The Log-Euclidean Fr$\acute{e}$chet mean $\overline{w}(\mathcal{B})$ of a batch $\mathcal{B}$ of SPD matrices $\{S_i\}_{i=1}^{|\mathcal{B}|}$ is defined as follows:
\begin{align}\label{eq:LEM_bar}
\overline{w}(\mathcal{B}) := \exp{\bigg(\frac{1}{|\mathcal{B}|} \sum_{i=1}^{|\mathcal{B}|} \log{(S_i)}\bigg)}.
\end{align}

Another key Riemannian metric for SPD manifolds is the affine-invariant Riemannian metric ($g^{AIRM}$)~\cite{pennec2006riemannian}, which we compare with the Log-Euclidean metric.
Formally, for any $P\in \mathcal{S}_{++}$, the $g^{AIRM}$ at $P$ is defined as:
\[
g^{AIRM}_P (v, w):= \langle \, P^{-\frac{1}{2}}\, v \,P^{-\frac{1}{2}}, P^{-\frac{1}{2}}\, w \,P^{-\frac{1}{2}} \, \rangle_{\mathcal{F}},
\]
where tangent vectors $v, w \in \mathcal{T}_{P} \mathcal{S}_{++}$. 
The manifold ($\mathcal{S}_{++}$, $g^{AIRM}$) preserves critical geometric properties, such as affine invariance, scale invariance, and inversion invariance~\cite{minh2022covariances}. These properties make this metric theoretically appealing for covariance-based learning tasks.

\subsection{Discrete Optimal Transport}~\label{Appen:Monge-Kantorovich}
Discrete optimal transport aims to find the minimum-cost transportation plan between source and target distributions that are accessible only through finite sample observations~\cite{kolouri2017optimal}.
Formally, given two measurable spaces $X$ and $\bar{X}$, the empirical distributions of measures $\mu \in X$ and $\nu \in  \bar{X}$ are expressed as $\mu:= \sum_{i=1}^N p_i \delta_{x_i}$ and $\nu:= \sum_{j=1}^M q_i \delta_{ \bar{x}_i}$, where the discrete variable $X$ takes values $\{x_i\}_{i=1}^N \in X$ with weights satisfying $\sum_{i=1}^N p_i =1$, while $ \bar{X}$ assumes values $\{\bar{x}_j\}_{j=1}^M \in \bar{X}$ with weights $\sum_{j=1}^M q_j =1$. 
The Dirac measure $\delta_{x}$ represents a unite point mass concentrated at the point $x \in X$ or $\bar{X}$. 
The Kantorovich formulation of the discrete optimal transport problem is defined as 
\[
\min_{\gamma \in \prod(\mu, \nu)} \big\langle\gamma, c(x, \bar{x})\big\rangle_{\mathcal{F}},
\]
where the transportation plan is constrained to the set $\gamma \in \prod(\mu, \nu):=\{\gamma \in \mathbb{R}_{ \geq 0 }^{|X| \times | \bar{X}|}  \big|  \gamma \mathbbm{1}_{|X|} = \mu, \text{ and } \gamma^\top \mathbbm{1}_{| \bar{X}|} = \nu\}$, and $\mathbbm{1}_d$ denotes a $d-$dimensional all-one vector.

\subsection{Joint Distribution Optimal Transport}
Assuming that the transport between source and target distributions can be estimated via optimal transport, Joint Distribution Optimal Transport (JDOT)~\cite{courty2017joint} extends the OT-DA framework by jointly optimizing both the optimal coupling $\gamma$ and the prediction function $f$. This is formulated as follows, 
\[
\min_{\gamma \in \prod(\mu, \nu)} \big\langle\gamma,c(x, y;\, \bar{x},f(\bar{x}))\big\rangle_{\mathcal{F}}, 
\]
where $y$ denotes the label set for samples $x$, $f$ is the prediction function, and the joint cost function combines a distance metric with a cross-entropy loss: $c(x, y; \bar{x}, f(\bar{x})):= \alpha_1 d_{\ell^2}(x, \bar{x}) + \alpha_2 \mathcal{L}(y, f(\bar{x}))$. DeepJDOT further enhances this framework by employing neural networks to embed raw data into a feature space. Its joint objective function becomes: $\mathcal{L}_{total}:= \mathcal{L}(y, f(g(x))) + \big\langle\gamma, c\big(g(x), y; \, g(\bar{x}), f(g(\bar{x})\big) \big\rangle_{\mathcal{F}}$, where $g$ represents the neural network encoder~\cite{damodaran2018deepjdot}.

\subsection{EEG-Based Motor Imagery Classifier on SPD Manifolds}

Extensive research has explored various classes of Riemannian metrics on SPD manifolds, with applications spanning neuroimaging and machine learning. These efforts have led to the development of manifold-valued statistical methods aimed at addressing computational challenges inherent to SPD manifolds—particularly in analyzing neuroimaging data with intrinsic geometric structures~\cite{pennec2006statistical,pennec2020manifold}. In the feature engineering era, Barachant et al. pioneered the use of Riemannian distances between EEG spatial covariance matrices for motor imagery classification, marking a major advancement in brain-computer interface research. Their method achieved strong performance in both multiclass classification and real-time applications~\cite{barachant2011multiclass,barachant2013classification,congedo2017riemannian}. Building on this foundation, Ju et al. advanced geometric deep learning methods on SPD manifolds in the deep learning era, developing architectures that extract discriminative spatio-temporal-frequency features~\cite{ju2020federated,ju2022tensor,ju2023graph,10340899,ju2024deep}. The neural network-based architecture in these works originated from SPDNet~\cite{huang2017riemannian}, a foundational architecture designed to process SPD matrices while preserving their manifold structure through the following layers:
\begin{itemize}
\item BiMap Layer: This layer transforms the covariance matrix $S$ by applying the BiMap transformation $WSW^T$, where the transformation matrix $W$ must have full row rank. 
\item ReEig Layer: This layer introduces non-linearity to SPD manifolds using $U\max{(\epsilon I, \Sigma)}U^T$, where $S = U\Sigma U^T$, $\epsilon$ is a rectification threshold, and $I$ is an identity matrix. 
\item LogEig Layer: This layer maps elements on SPD manifolds to their tangent space using $U \log{(\Sigma)} U^T$, where $S = U \Sigma U^T$. The logarithm function applied to a diagonal matrix takes the logarithm of each diagonal element individually.
\end{itemize}

\section{Methodology}~\label{methodology}

In this section, we will generalize the OT-DA framework on ($\mathcal{S}_{++}$, $g^{LEM}$) and propose a novel geometric deep learning approach to address the problems formulated on this framework. 

\subsection{OT-DA Framework on ($\mathcal{S}_{++}$, $g^{LEM}$)}
we first calculate the gradient of any squared distance function on general Riemannian manifolds. 

\begin{lemma}~\label{l1}
Let $(\mathcal{M}, g)$ be a connected, compact, and $C^3-$smooth Riemannian manifold without a boundary. Consider a compact subset $\mathcal{U} \subset \mathcal{M}$ and a fixed point $q \in \mathcal{U}$. Let $f(p) = \frac{1}{2} d_g^2(p, q)$ denote the squared distance function. Then, for $p$ not on the boundary of $\mathcal{M}$, the gradient of $f(p)$ is given by $\nabla f(p) = -\log_p(q)$.
\end{lemma}

\begin{proof}[Proof]
Without loss of generality, consider a point $p \in \mathcal{M}^{\circ}$. There exists an open neighborhood $\mathcal{N}$ of $p$ and $\epsilon >0$ such that the exponential map $\exp_p$ maps the ball $\mathcal{B}(0, \epsilon) \subset T_p\mathcal{M}$ diffeomorphically onto $\mathcal{N}$~\cite[Corollary 5.5.2.]{petersen2006riemannian}. Let $c(s)$ be a geodesic on $\mathcal{N}$, parameterized with constant speed, with $c(0) := q \in \mathcal{N}$ and $c(1) := p$ with $\dot{c}(0) = \log_q(p)$. By reversing the direction of the geodesic $c(1-s)$, we have $\dot{c}(1) = -\log_p(q)$. We proceed by linearizing the exponential maps around $0 \in T_p\mathcal{M}$ and $\dot{c}(0)\in T_q\mathcal{M}$ and derive the derivative of a function $f$ as follows:
\begin{align*}
& f\big(\exp_p (v)\big) \\
=\, &\frac{1}{2} \cdot d_g^2\big(\exp_p (v), q\big)\\
=\, &\frac{1}{2} \cdot \big|\log_q \big(\exp_p (v)\big)\big|_q^2\\
=\, & \frac{1}{2} \cdot \big|\dot{c}(0) + (D\log_q)_{\dot{c}(0)} \big((D\exp_p)_0 v\big) + o(|v|_p)\big|_q^2\\
=\, &\frac{1}{2} \cdot \big|\dot{c}(0)\big|_q^2 + g_q(\dot{c}(0), (D\log_q)_{\dot{c}(0)} v) + o(|v|_p) \\
=\, &\frac{1}{2} \cdot d_g^2(p, q) + g_p(\dot{c}(1), v) + o(|v|_p),
\end{align*}
where the differential $(D\exp_p)_q(\cdot): T_q (T_p \mathcal{M}) \longmapsto (T_p \mathcal{M})$ is the differential of the exponential map at $q$ with respect to $p$. The fourth equality follows from $(D\exp_p)_0 = I$ on $T_p\mathcal{M}$, and the fifth equality follows from Gauss' lemma~\cite[Section 3.3.5.]{do1992riemannian}. Hence, by the definition of the gradient on Riemannian manifolds, we have $df(p)(v) = g_p\big(\nabla f(x), v\big)$, which implies that $\nabla f(p) = \dot{c}(1) = -\log_p(q)$.
\end{proof}

We say a function $\psi: \mathcal{M}\longmapsto \mathbb{R}\cup\{-\infty\}$ is \emph{c-concave} if it is not identically $-\infty$ and there exists $\varphi: \mathcal{M}\longmapsto \mathbb{R}\cup\{\pm \infty\}$ such that
\[
\psi(p) = \inf_{q \in \mathcal{M}} \big\{ c(p, q) + \varphi(q) \big\},
\]
where $c(p,q)$ is a nonnegative cost function that measures the cost of transporting mass from $q$ to $p \in \mathcal{M}$. 
Brenier's polar factorization on general Riemannian manifolds is given as follows:
\begin{lemma}[Brenier's Polar Factorization~\cite{mccann2001polar}]~\label{l2}
Given any compactly supported measures $\mu, \nu \in \mathbb{P}(\mathcal{M})$. If $\mu$ is absolutely continuous with respect to Riemannian volume, there exists a unique optimal transport map $T \in S(\mu, \nu)$ as follows, 
\[
T(x) = \exp_p \big(-\nabla \psi (p)\big), 
\]
where $\psi:\mathcal{M}\longmapsto \mathbb{R} \cup \{\pm \infty\}$ is a c-concave function with respect to the cost function $c(p, q) = \frac{1}{2} \cdot d_g^2(p, q)$, and $\exp_p(\cdot)$ is the exponential map at $p$ on $(\mathcal{M}, g)$. 
\end{lemma}

On the SPD manifold ($\mathcal{S}_{++}$, $g^{LEM}$), let's consider source measure $\mu$ and target measure $\nu \in \mathbb{P}(\mathcal{S}_{++})$ that are accessible only through finite samples $\{S_i\}_{i=1}^N \sim \mu$ and $\{\bar{S}_j\}_{j=1}^M \sim \nu$. 
Each sample contributes to its respective empirical distribution with weights $\frac{1}{N}$ and $\frac{1}{M}$.
We assume there exists a linear transformation on ($\mathcal{S}_{++}$, $g^{LEM}$) between two discrete feature space distributions of the source and target domains that can be estimated with the discrete Kantorovich problem, i.e., there exists
\begin{equation}~\label{OT_2}
\gamma^{\dagger}:= \arg\min_{\gamma \in \prod(\mu, \nu)} \big\langle\gamma, d_{g^{LEM}}^2(S, \bar{S})\big\rangle_{\mathcal{F}}, 
\end{equation}
where transportation plan is as follows,
\[
\prod(\mu, \nu):=\{\gamma \in \mathbb{R}_{ \geq 0 }^{N \times M} \big| \gamma \mathbbm{1}_{N} = \frac{1}{N}, \text{ and } \gamma^\top \mathbbm{1}_{M} = \frac{1}{M}\}.
\]

The pseudocode for the computation of the transport plan is presented in Algorithm~\ref{alg}.

\begin{algorithm}[!h]
\KwInput{Source samples $\{S_i\}_{i=1}^N \sim \mu$ and target samples $\{\bar{S}_j\}_{j=1}^M \sim \nu$.
 }
 \KwOutput{Transport plan $\gamma$ and new coordinates for target samples $\{\bar{S}^{\dagger}_j\}_{j=1}^M$.}
1. Compute ground metric matrix for OT as follows:\\
\hspace*{7.5em} $D(i, j) \gets d_{g^{LEM}}(S_i, \bar{S}_j)$.\\
2. Compute transport plan $\gamma$ using the earth movers distance-based OT algorithm as follows:
\hspace*{7em} $\gamma^{\dagger} \gets OT^{EMD}(N, M, D(i,j))$.\\
3. Compute new coordinates using transport plan on ($\mathcal{S}_{++}$, $g^{LEM}$) as follows:
\hspace*{7em} $\{\bar{S}^{\dagger}_j\}_{j=1}^M \gets \exp{\big(\gamma^{\dagger}[j,:] \cdot \{ \log (S_i) \}_{i=1}^N \big)}$.
\caption{Discrete Monge-Kantorovich Algorithm on ($\mathcal{S}_{++}$, $g^{LEM}$).~\label{alg}}
\end{algorithm}

In the following paragraph, we will employ discrete transport plans to construct c-concave functions and utilize Lemma~\ref{l1} and~\ref{l2} to obtain continuous transport plans.
According to Algorithm~\ref{alg}, the new coordinates for target samples on SPD manifolds are given by the following expression:
\[
\{\bar{S}^{\dagger}_j\}_{j=1}^M := \exp{\big(\gamma^{\dagger}[j,:] \cdot \{ \log (S_i) \}_{i=1}^N \big)}.
\]
We define $\varphi(\bar{S}):= + \infty$ for $\bar{S} \notin \{\bar{S}^{\dagger}_j\}_{j=1}^M$, and set the cost function $c(S, \bar{S}) := \frac{1}{2}\cdot d_{g^{LEM}}^2(S, \bar{S})$.
This leads to the formulation of the corresponding c-concave function as follows:
\[
\psi(S) = \min_{j \in \{1, ..., M\}} \frac{1}{2}\cdot d_{g^{LEM}}^2(S, \bar{S}^{\dagger}_j) + \varphi(\bar{S}^{\dagger}_j).
\]
For a fixed $S$, we suppose $j^\ast  \in \{1, ..., M\}$ achieves the minimum. By invoking Lemma~\ref{l1} and~\ref{l2}, we promptly derive the unique optimal transport map as follows,
\[
T(S) = \exp_{S} \big(-\nabla_S \psi (S) \big) = \bar{S}^{\dagger}_{j^\ast} \in \{\bar{S}^{\dagger}_j\}_{j=1}^M, 
\]
which implies the following theorem:
\begin{theorem}~\label{t1}
On ($\mathcal{S}_{++}$, $g^{LEM}$), consider the source measure $\mu$ and target measure $\nu \in \mathbb{P}(\mathcal{S}_{++})$, which are only accessible through finite samples $\{S_i\}_{i=1}^N \sim \mu$ and $\{\bar{S}_j\}_{j=1}^M \sim \nu$. Each sample contributes to the empirical distribution with weights $\frac{1}{N}$ and $\frac{1}{M}$, respectively.
Assuming the existence of a linear transformation between the feature space distributions of the source and target domains, it can be estimated using the discrete Monge-Kantorovich Problem~\ref{OT_2}.
Therefore, the values of the continuous optimal transport $T(S)$ lie in the set $\big\{ \exp{\big(\gamma^{\dagger}[j,:] \cdot { \log (S_i) }_{i=1}^N \big)} \big\}_{j=1}^M$.
\end{theorem}

\begin{remark}

1). Theorem~\ref{t1} extends \cite[Theorem 3.1]{courty2016optimal} to the OT-DA framework on SPD manifolds equipped with the Log-Euclidean metric. Unlike conventional OT-DA methods restricted to affine transformations, our framework uniquely incorporates arbitrary supervised neural network-based mappings while minimizing transportation costs via the squared Riemannian distance $d_{g^{LEM}}^2(\cdot,\cdot)$.

2). In Appendix~\ref{appen:euclidean_cost}, we contrast our approach with the SPD manifold OT-DA framework in~\cite{yair2019domain}, which employs the squared Euclidean distance as its cost function. This comparison highlights the geometric advantages of our Riemannian distance formulation over Euclidean approximations.

\end{remark}

\begin{figure*}[!h]
\centering
\includegraphics[width=\linewidth]{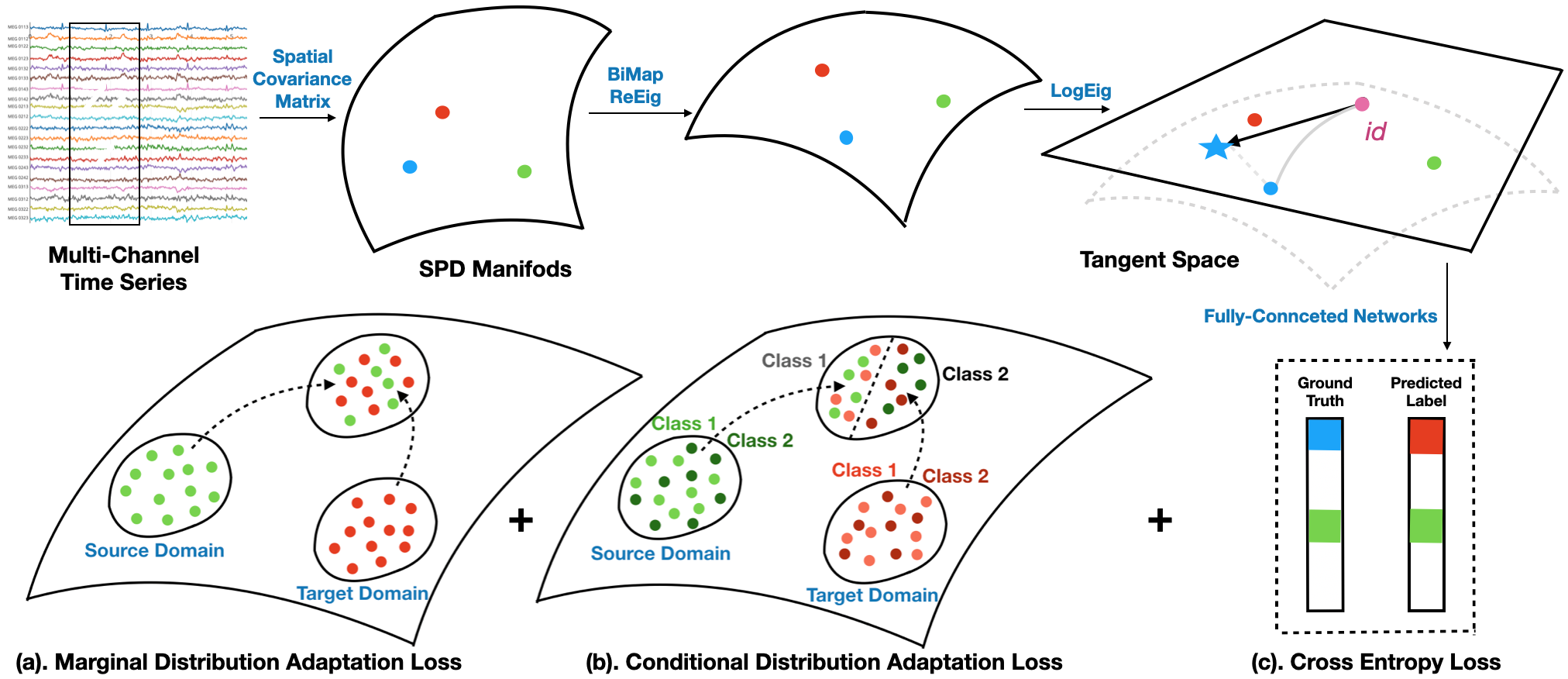}
\caption{Illustration of Deep Optimal Transport: Multi-channel signals are first transformed into covariance matrices, situating these matrices on the space of ($\mathcal{S}_{++}$, $g^{LEM}$). The SPD matrix-valued data is processed through the BiMap, ReEig, and LogEig layers, moving it to the tangent space at the identity. The loss function takes into account the cross-entropy loss, marginal distribution adaptation, and conditional distribution adaptation.
\label{fig:DOT}
} 
\end{figure*}

\subsection{Deep Optimal Transport}~\label{sec:DOT_SPD}
We propose Deep Optimal Transport (DOT), a SPDNet-based architecture on ($\mathcal{S}_{++}$, $g^{LEM}$), to address joint distribution adaptation under both marginal and conditional shifts. Unlike traditional OT methods, DOT leverages a deep encoder to learn nonlinear push-forward mappings that align source and target distributions by minimizing a composite OT loss. The key assumption is that domain discrepancies can be bridged through transport plans parameterized by neural networks. As illustrated in Figure~\ref{fig:DOT}, the architecture integrates:

\begin{enumerate}[i.]
\item Nonlinear Manifold Mapping $\varphi: X_S, X_T \longmapsto Z$: Using multiple BiMap layers and ReEig layers from SPDNet as transformations on SPD manifolds.
\item Projection: Applying the LogEig layer from SPDNet to flatten the SPD manifold-valued data onto the Euclidean tangent space at identity.
\item Joint OT-Driven Distribution Alignment: Simultaneously minimizing the OT costs for marginal and conditional distributions as follows:

\subsubsection*{Marginal Distribution Adaptation (MDA)}

The primary objective of MDA is to align the marginal distributions $\mathbb{P}_S\big(\varphi(X_S)\big)$ and $\mathbb{P}_T\big(\varphi(X_T)\big)$ on the Riemannian manifold ($\mathcal{S}_{++}$, $g^{LEM}$). 
We quantify their statistical discrepancy through the Riemannian distance between their Log-Euclidean Fréchet means. Specifically, for batches of source and target samples $\mathcal{B}_S$ and $\mathcal{B}_T$, the MDA loss is defined as the Frobenius norm of the difference between their logarithmic mean matrices:
\[
\mathcal{L}_{MDA} := || \log \big(\overline{w}(\mathcal{B}_S)\big) - \log \big(\overline{w}(\mathcal{B}_T) \big)||_{\mathcal{F}}.
\]

\subsubsection*{Conditional Distribution Adaptation (CDA)}

CDA aims to match the class-conditional distributions $\mathbb{Q}_S(\varphi(Y_S)\big|\varphi(X_S))$ and $\mathbb{Q}_T(\varphi(Y_T)\big|\varphi(X_T))$ on the same manifold ($\mathcal{S}_{++}$, $g^{LEM}$), contrasting with traditional posterior probability adaptation. In unsupervised settings, pseudo-labels $\hat{y}$ for target domain samples (generated by a baseline algorithm) are utilized to construct class-specific sample batches. For each class $\ell \in \{1, ...., L\}$, we define $S^{\ell}:=\{x_S |\hat{y}_S={\ell}\}$ and $T^{\ell} :=\{x_T |\hat{y}_T=\ell\}$, then compute the Riemannian distance between their Fréchet means. The CDA loss is formulated as: 
\[
\mathcal{L}_{CDA} := ||\log \big(\overline{w}(\mathcal{B}_{S^{\ell}}) \big)- \log \big(\overline{w}(\mathcal{B}_{T^{\ell}}) \big)||_{\mathcal{F}}.
\]

\subsubsection*{Overall Objective Function}\label{Correction Loss}

The overall objective function of the DOT framework combines the cross-entropy loss $\mathcal{L}_{CE}$ with joint distribution adaptation terms, formulated as:
\[
\mathcal{L}_{DOT}:= \alpha_1 \mathcal{L}_{CE} + \alpha_2 \mathcal{L}_{MDA}^2 + \alpha_3 \mathcal{L}_{CDA}^2,
\] 
where $\alpha_1, \alpha_2$, and $\alpha_3 \geq 0$ are weighting coefficients governing the trade-off between classification accuracy and distribution alignment.

\end{enumerate}

\begin{remark}

1). Based on Brenier's theorem and related theoretical results~\cite{mccann2001polar}, both $\mathcal{L}_{MDA}$ and $ \mathcal{L}_{CDA}$ adopt squared-distance formulations.

2). While multiple Riemannian metrics exist on $\mathcal{S}_{++}$, the affine invariant Riemannian metric (AIRM) $g^{AIRM}$ is unsuitable for our framework. Computing the Fr$\acute{e}$chet mean under ($\mathcal{S}_{++}$, $g^{AIRM}$) requires iterative optimization, which is incompatible with deep learning loss functions requiring gradient computation. 

3). By substituting Equation~\ref{eq:LEM_bar} into $\mathcal{L}_{MDA}$ and $\mathcal{L}_{CDA}$, we obtain:
\begin{align*}
\mathcal{L}_{MDA} = || \sum_{S_i \in \mathcal{B}_S} \frac{\log{(S_i)}}{|\mathcal{B}_S|} - \sum_{S_j \in \mathcal{B}_T} \frac{\log{(S_j)}}{|\mathcal{B}_T|}||_{\mathcal{F}}; \hspace*{1em} \mathcal{L}_{CDA} = || \sum_{{S^{\ell}_i} \in \mathcal{B}_{S^l}} \frac{\log{({S^{\ell}_i})}}{|\mathcal{B}_{S^{\ell}}|} - \sum_{{S^{\ell}_j} \in \mathcal{B}_{T^{\ell}}} \frac{\log{({S^{\ell}_j})}}{|\mathcal{B}_{T^{\ell}}|}||_{\mathcal{F}}.
\end{align*}
These simplified forms have dual interpretations: 
\begin{itemize}
\item Empirical Maximum Mean Discrepancy (MMD): Let $n = |\mathcal{B}_S|$, $m = |\mathcal{B}_T|$, $\varphi{(x)}:= \log{(x)}$, and $\mathcal{H}:= \mathbb{R}^d$. The loss becomes $\mathcal{L}_{MMD} = ||\sum_{i=1}^n \varphi{(S_i)}/n - \sum_{j=1}^m \varphi{(S_i)}/m||_{\mathcal{H}}$.
\item Correlation Alignment (CORAL): As covariance matrices of multi-channel signals, the alignment of second-order statistics minimizes domain drift~\cite{sun2016deep,sun2017correlation}.
\end{itemize}

4). Multi-Source Adaptation: For multi-source scenarios, we extend the loss by integrating pairwise source-target adaptations on ($\mathcal{S}_{++}$, $g^{LEM}$):
\[
\mathcal{L}:= \alpha_1^i \mathcal{L}_{CE} + \sum_i \Big( \alpha_2^i \mathcal{L}_{MDA, i}  + \alpha_3^i \mathcal{L}_{CDA, i} \Big),
\]
where $i^{th}$ indexes source domains. This aligns with multi-marginal optimal transport theory on manifolds~\cite{kim2015multi} and multi-source joint transport~\cite{turrisi2022multi}.

5). DOT Framework: DOT does not explicitly solve an OT problem. Instead, it leverages transport costs as a guiding principle for neural network training. The trained network maps source and target domains onto a submanifold of the SPD manifold, effectively implementing a transfer learning subspace method.

6). Adaptation Strategy: Unlike DeepJDOT, which computes transport costs for all points and requires alternating parameter updates due to its deterministic transport plan, DOT calculates costs only for centroids and achieves conditional marginal adaptation without alternating updates.

7). Perspective from Bures-Wasserstein Geometry: Bures-Wasserstein Geometry is a geometric framework on SPD manifolds, where the 2-Wasserstein distance between centered Gaussian distributions induces the Bures metric, governing optimal transport and barycenter computations for covariance matrices~\cite{takatsu2011wasserstein}. Specifically, for centered Gaussian distributions $\mathcal{N}(0, C_s)$ and $\mathcal{N}(0, C_t)$, the 2-Wasserstein distance reduces to the Bures metric on the SPD manifold:
\[
W_2^2\big(\mathcal{N}(0, C_s), \mathcal{N}(0, C_t)\big) = \text{Tr}\left(C_s + C_t - 2\left(C_s^{1/2}C_t C_s^{1/2}\right)^{1/2}\right).
\]
The barycenter of Gaussians $\{\mathcal{N}(0, C_i)\}_{i=1}^n$ under $W_2$ is defined via their covariances:
\[
\bar{C} = \arg\min_{C \succ 0} \sum_{i=1}^n W_2^2\big(\mathcal{N}(0, C), \mathcal{N}(0, C_i)\big),
\]
with efficient computation via fixed-point iterations~\cite{agueh2011barycenters,alvarez2016fixed}. 
When the deep encoder $\varphi$ is identity (i.e., $C_i = x_i x_i^T$) and data follows centered Gaussians with commuting covariances ($C_s C_t = C_t C_s$), $\mathcal{L}_{MDA}$ under $d_{\text{LEM}}$ recovers the Bures metric. 
This is because the logarithmic transformation under $g^{LEM}$ aligns with the geometric mean in Bures for commuting matrices. 
However, the DOT framework extends beyond this idealized scenario in two ways: 
\begin{enumerate}[i.]
\item The deep encoder $\varphi$ enables nonlinear manifold mapping, breaking the Gaussian assumption; 
\item The composite loss $\mathcal{L}_{\text{CDA}}$ jointly aligns conditionals, which is nontrivial under pure Wasserstein barycenter frameworks.
\end{enumerate}
\end{remark}

\section{EXPERIMENTS}~\label{experiments}

\subsection{Deep Optimal Transport-Based Motor Imagery Classifier}~\label{sec:DOT_BCI}

\begin{figure}[!h]
\centering
\includegraphics[width=0.6\linewidth]{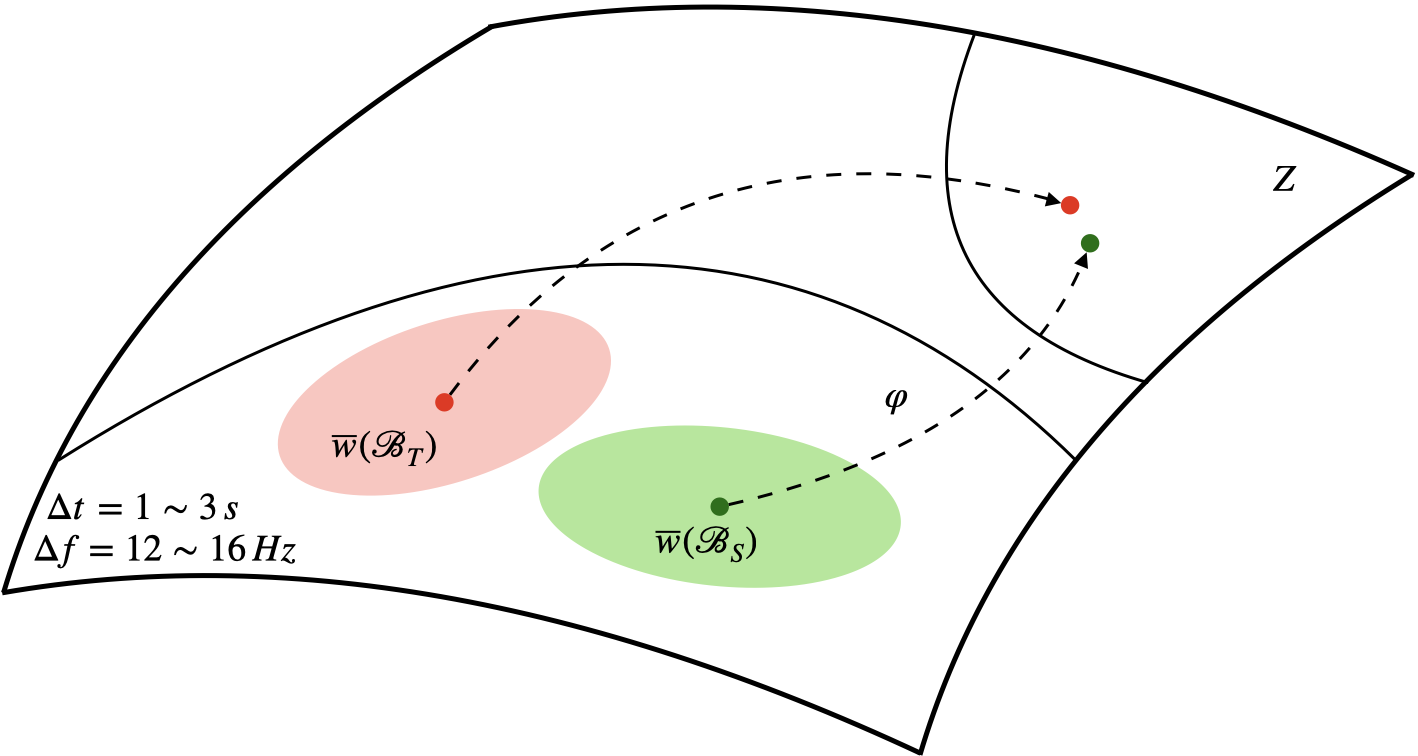}
\caption{Illustration of DOT-Based Motor-Imagery Classifier: 
The proposed neural networks $\varphi$ simultaneously transports the Fréchet means $\overline{w}(\mathcal{B}_{S})$ and $\overline{w}(\mathcal{B}_{T}$ of each batch of source and target samples within each frequency band to a common subspace $Z$ associated with that frequency band. Assumption~\ref{a2} guarantees that all frequency components are equally weighted in the classification process. Assumption~\ref{a3} ensures that the transformations applied to the source and target domains are restricted to occur within corresponding EEG frequency bands.
\label{fig:channel}
} 
\end{figure}

We model EEG spatial covariance matrices on the Riemannian manifold ($\mathcal{S}_{++}$, $g^{LEM}$) and employ DOT to address statistical distribution shifts during feedback phases.
Motor imagery classification leverages event-related desynchronization/synchronization effects~\cite{pfurtscheller1997eeg}, which identify discriminative EEG frequency components in specific bands.
The BiMap layer projects source and target features into a shared latent space where $X_S \approx X_T$, enabling SPD manifold modeling with the Log-Euclidean metric. 
This layer induces a transitive group action on the manifold, allowing smooth transitions between any two SPD points (Fig.~\ref{fig:channel} illustrates the DOT-based classifier architecture).

Formally, an EEG trial is represented by a matrix $X \in \mathbb{R}^{n_C \times n_T}$, where $n_C$ denotes the number of channels and $n_T$ the number of time points.
To capture localized spatio-spectral dynamics, we compute spatial covariance matrices $S(\Delta f, \Delta t) \in \mathcal{S}_{++}^{n_C}$, where each matrix characterizes the signal within a specific frequency band $\Delta f$ and time interval $\Delta t$.
Let $\mathcal{F}$ and $\mathcal{T}$ denote the overall frequency and time domains, respectively. A segmentation over these domains is defined as
\[
\text{segment}(\mathcal{F},\mathcal{T}):=\big\{ (\Delta f \times \Delta t) \, \big| \, \Delta f \subseteq \mathcal{F}, \Delta t \subseteq \mathcal{T}\big\}.
\]
To ensure practical convergence in real-world scenarios, the segmentation must satisfy the following conditions:

\begin{assumption}[Necessary Optimal Transport Condition]~\label{a1}
There exist linear transformations between the feature space distributions on the source and target domains for each segment of $\text{segment}(\mathcal{F}, \mathcal{T})$, which can be estimated using the Kantorovich Problem~\ref{OT_2}.
\end{assumption}

\begin{assumption}[Uniform Spectral Prior]~\label{a2}
The weights for the Log-Euclidean Fr$\acute{e}$chet mean of samples on each time-frequency segment $\Delta f \times \Delta t \in \text{segment}(\mathcal{F},\mathcal{T})$ in both the source and target empirical distributions are uniform and equal $1/|\text{segment}(\mathcal{F},\mathcal{T})|$.
\end{assumption}

\begin{assumption}[Feature Disjointness]~\label{a3}
For any time-frequency segment $\Delta f \times \Delta t \in \text{segment}(\mathcal{F}, \mathcal{T})$, let $\mathcal{B}_S$ and $\mathcal{B}_T$ denote corresponding batches of source and target domain samples, respectively. We require that the Riemannian distance between the Log-Euclidean Fr$\acute{e}$chet means of source and target samples, both under the same time-frequency segment, is minimized, that is:
\begin{align*}
d_{g^{LEM}}\big(\overline{w}(\mathcal{B}_{S}), \overline{w}(\mathcal{B}_{T})\big) &\leq d_{g^{LEM}}\big(\overline{w}(\mathcal{B}_{S}), \overline{w}(\mathcal{B}_{T}')\big),
\end{align*}
where $\mathcal{B}_T'$ is a batch of target samples obtained from a different segment $\Delta f' \times \Delta t' \in \text{segment}(\mathcal{F}, \mathcal{T})$, such that $\Delta f' \neq \Delta f$ or $\Delta t' \neq \Delta t$.

\end{assumption}

\begin{remark}

1). Assumption~\ref{a1} establishes the foundational requirements for formulating the OT-DA problem on Riemannian manifolds.

2). To address uncertainty in EEG frequency band weighting for motor imagery tasks, Assumption~\ref{a2} imposes a uniform prior across all spectral components. This implies equal contribution weights for all frequency bands in the classification objective, unless task-specific evidence dictates otherwise.

3). Assumption~\ref{a3} enforces non-overlapping feature distributions across distinct frequency bands, ensuring orthogonality in the latent representations learned from different spectral regions.

\end{remark}

\begin{corollary}~\label{c1}
Suppose EEG signals in the source and target domain have the same segmentation, \text{segment}($\mathcal{F},\mathcal{T}$). Assuming we model EEG spatial covariance matrices in ($\mathcal{S}_{++}^{n_C}$, $g^{LEM}$) and that Assumption~\ref{a1}, ~\ref{a2}, and~\ref{a3} exist, then the values of the continuous optimal transport $T(S)$ reside in the set $\big\{ \log (S_i) \big\}_{i=1}^N$.
\end{corollary}

\begin{proof}
In this case, the transport plan $\gamma$ equals the identity matrix. Consequently, Theorem~\ref{t1} leads directly to this corollary.
\end{proof}

\subsection{Evaluation Dataset}\label{dataset}
We evaluate our method on three publicly available EEG motor imagery datasets—Korea University (KU), BNCI2014001 (BCIC-IV-2a), and BNCI2015001—accessed via the MOABB package (\url{https://github.com/NeuroTechX/moabb}) and BNCI Horizon 2020 project (\url{http://bnci-horizon-2020.eu/database/data-sets}). All datasets include two-session recordings per subject (collected on different days) and were preprocessed using Chebyshev Type II filters with a 4 Hz bandwidth, 3 dB passband ripple, and 30 dB stopband attenuation. The descriptions of the three datasets are as follows:

1). Korea University (KU) Dataset: Also referred to as Lee2019\_MI, this dataset contains EEG recordings from 54 subjects performing a binary-class motor imagery task. EEG signals were acquired using 62 electrodes at a sampling rate of 1,000 Hz. For analysis, we focused on 20 electrodes located over the motor cortex. Each subject participated in two sessions (S1 and S2), each consisting of training and testing phases with 100 trials per phase, equally split between left- and right-hand imagery tasks. EEG epochs were extracted from 1 to 3.5 seconds post-stimulus, yielding a segment duration of 2.5 seconds.

2). BNCI2014001 Dataset: Also known as BCIC-IV-2a, this dataset comprises EEG signals from 9 subjects performing a four-class MI task (left hand, right hand, feet, tongue). EEG was recorded from 22 Ag/AgCl electrodes and three EOG channels at a sampling rate of 250 Hz. Signals were bandpass filtered between 0.5$\sim$100 Hz and notch filtered at 50 Hz. Each subject completed two sessions on different days, with 288 trials per subject, divided into six runs of 12 cue-based trials per class. Epochs were segmented from 2 to 6 seconds post-cue, resulting in a 4-second window.

3). BNCI2015001 Dataset: This dataset features 12 participants engaged in a binary MI task (right hand vs. both feet). Most participants completed two sessions on consecutive days, each comprising 100 trials per class (200 trials in total). Four participants also completed a third session, which was included in our evaluation. EEG was recorded at a sampling rate of 512 Hz and bandpass filtered between 0.5$\sim$100 Hz with a 50 Hz notch filter. Epochs were extracted from 3 to 8 seconds post-prompt, yielding a 5-second duration.

\subsection{Experimental Settings}
\begin{figure}[!h]
\centering
\includegraphics[width=0.5\linewidth]{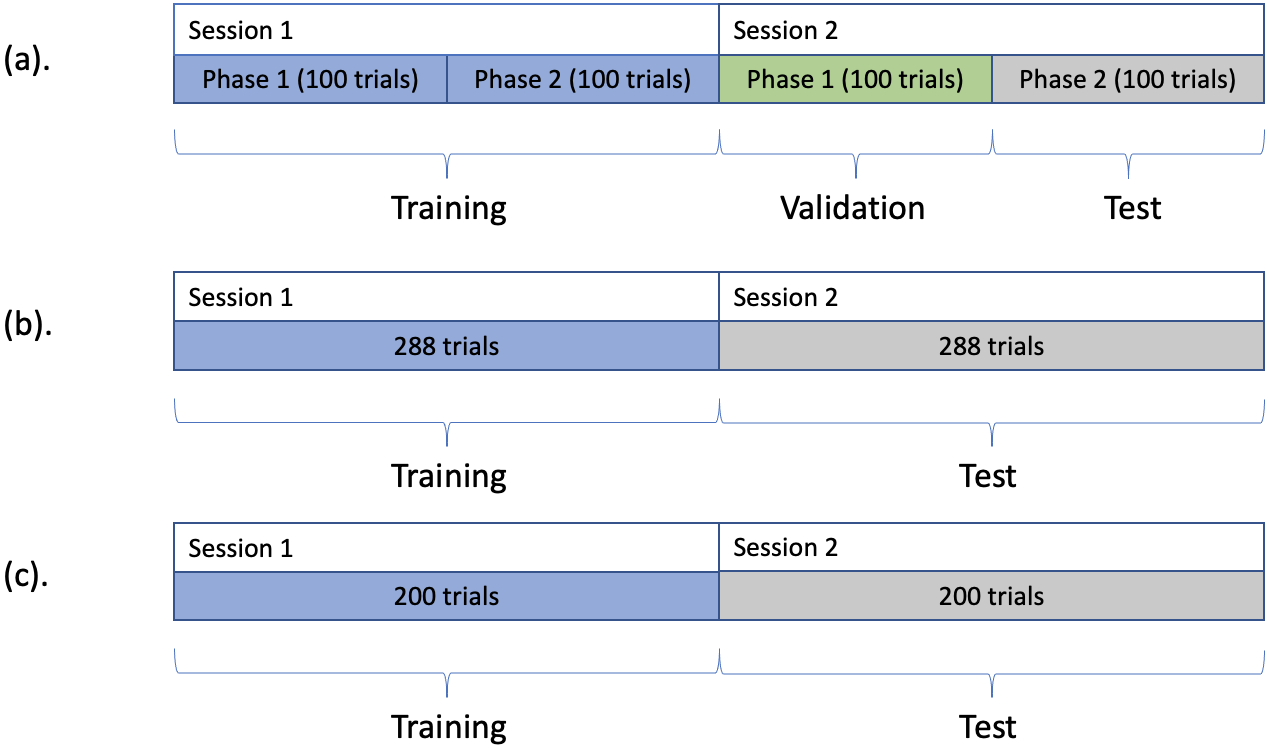}
\caption{Illustrations for experimental settings on three datasets:  (a). KU; (b). BNCI2014001; (c). BNCI2015001
\label{fig:exp_settings}
} 
\end{figure}

For domain adaptation, experiments followed two protocols, illustrated in Fig.~\ref{fig:exp_settings}:
1). Semi-supervised Domain Adaptation: For the KU dataset, S1 was used for training, and the first half of S2 was used for validation, while the second half was reserved for testing. Early stopping was adopted during the validation phase.
2). Unsupervised Domain Adaptation: For the BNCI2014001 and BNCI2015001 datasets, S1 was used for training, and S2 was used for testing. No validation set was used, and a maximum epoch value was set for stopping in practice.

We specified that the transfer is strictly unidirectional, i.e., from the source domain to the target domain $S1 \rightarrow S2$, reflecting real-world calibration-to-feedback workflows and avoiding counterproductive negative transfer.

\subsection{Evaluated Baselines}

The geometric methods employed in this study can be broadly categorized into four groups: parallel transport, deep parallel transport, optimal transport, and deep optimal transport. Each category encompasses distinct transport methodologies that are systematically applied to three fundamental machine learning classifiers: Support Vector Machines (SVM), Minimum Distance to Riemannian Mean (MDRM), and SPDNet. We first present an overview of these base classifiers before detailing the transport methods within each methodological category.

\begin{enumerate}
\item Support Vector Machine (SVM)~\cite{cortes1995support}: This is a well-known machine learning classifier used for classification and regression tasks, which aims to find an optimal hyperplane that maximally separates data points of different classes, allowing it to handle linear and non-linear classification problems effectively. 
\item Minimum Distance to Riemannian Mean (MDRM)~\cite{barachant2011multiclass}: This classifier employs the geodesic distance on SPD manifolds for classification. More specifically, a centroid is estimated for each predefined class, and subsequently, the classification of a new data point is determined based on its proximity to the nearest centroid.
\item SPDNet~\cite{huang2017riemannian}: This classifier was described in the previous section. In the experiments, we utilize a single-layer BiMap architecture where the input and output dimensions align with the number of electrodes in the primary motor cortex: 20 (KU dataset), 22 (BNCI2014001 dataset), and 13 (BNCI2015001 dataset).
\end{enumerate}

Since all computations on the SPD manifold inherently involve Riemannian metrics, we explicitly identify the Riemannian metric associated with each method. The variations in transport methods across methodological categories, distinguished by their underlying Riemannian metrics, are systematically organized in the first three columns of Table~\ref{tab:comparative_experiments}.

\begin{enumerate}[i.]
\item Category of Parallel Transport (PT): This category encompasses methods that apply parallel transport to align training and test data distributions across domains. A prominent technique in this framework is Riemannian Procrustes Analysis (RPA)~\cite{rodrigues2018riemannian}, which operates on the SPD manifold equipped with the Affine-Invariant Riemannian Metric. RPA employs three sequential geometric transformations: 
\begin{enumerate}[(1).]
\item Recentering (RCT): Aligns dataset centroids to the identity matrix via parallel transport from the Fréchet mean (of either source or target domain) to $I$. 
\item Stretching: Rescales data distributions to equalize dispersion around the mean. 
\item Rotation (ROT): Adjusts the orientation of the target domain’s point cloud to match the source domain. 
\end{enumerate}
Notably, ROT functions as an unsupervised domain adaptation step, while RCT is restricted to semi-supervised settings due to its dependency on labeled source data. The RCT method, originally proposed as an independent solution for cross-session motor imagery classification~\cite{yair2019parallel}, is integrated here as the foundational alignment step in RPA. Post-transformation, classification is performed using the MDRM algorithm.

\item Category of Deep Parallel Transport (DPT): This framework integrates deep learning with parallel transport operations on the SPD manifold. A key implementation is Riemannian Batch Normalization (RieBN)~\cite{brooks2019riemannian}, an SPD-valued neural network architecture that leverages parallel transport for batch normalization. RieBN performs intrinsic centering and bias correction directly on the manifold, ensuring geometrically consistent feature distributions across layers.

\item Category of Optimal Transport (OT): These methods formulate cross-session covariance matrix distribution alignment as an OT problem, solved using Wasserstein distance variants. The classifier is trained on session-specific data and evaluated under session-shifted conditions, operating in a fully unsupervised regime. Three Wasserstein metrics~\cite{bonet2023sliced} are applied:
\begin{enumerate}[(1).]
\item Earth Mover’s Distance (EMD);
\item Sliced Wasserstein Distance on SPD matrices (SPDSW);
\item Log-Euclidean Sliced Wasserstein Distance (logSW).
\end{enumerate}

\item Category of Deep Optimal Transport (DOT): The proposed method unifies SPDNet and OT on the SPD manifold equipped with Log-Euclidean Metric. By jointly optimizing a transport plan and a deep feature extractor, it projects source and target domains into an OT-aligned latent space, enabling invariant classification under domain shifts.

\end{enumerate}

\section{Main Results}\label{sec:main_results}

\begin{figure}[!h]
\center
\includegraphics[width=0.9\linewidth]{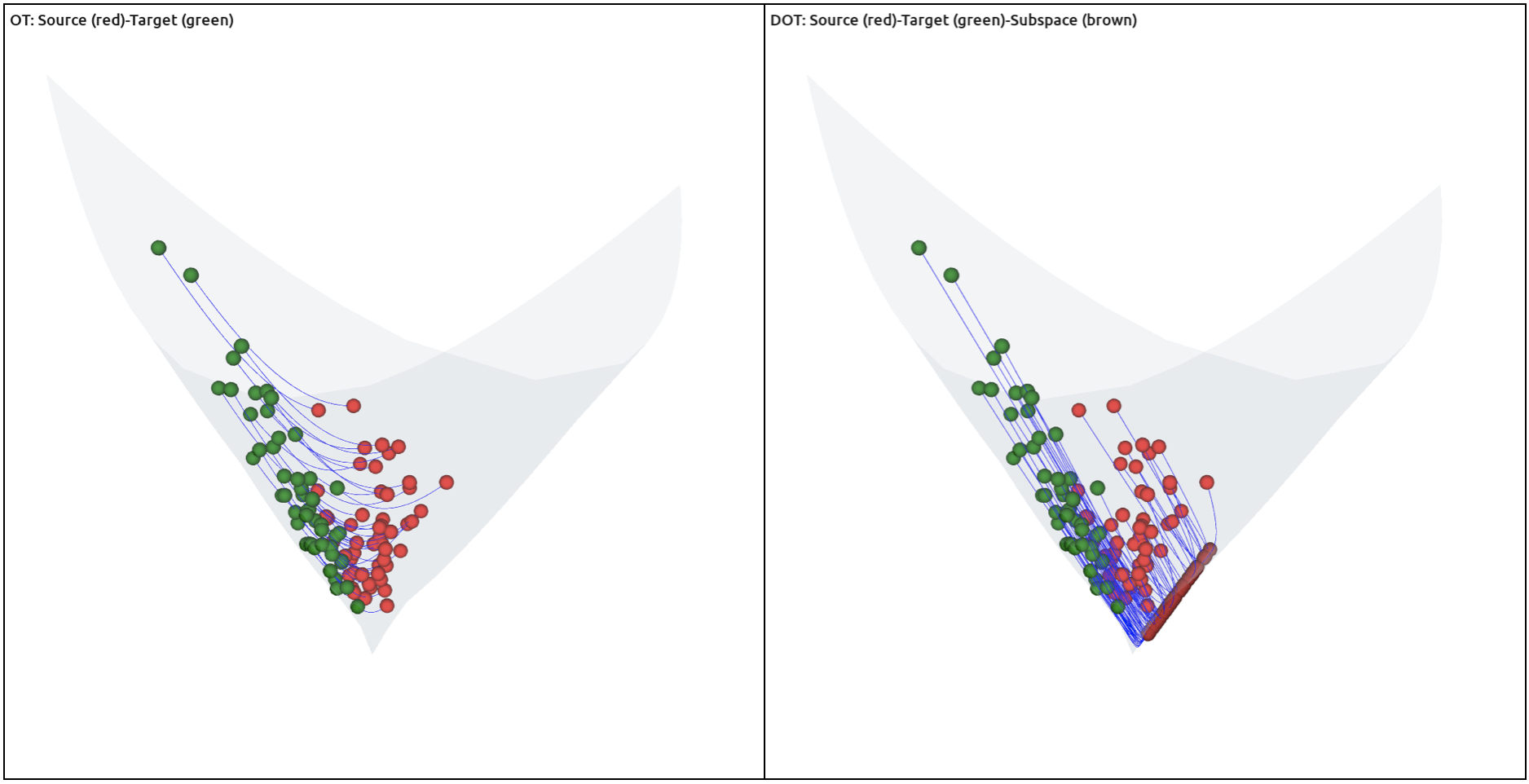}
\caption{Comparison of OT and DOT on 2 by 2 SPD matrices: Each colored ball represents a 2 by 2 SPD matrix. The left diagram illustrates the OT method, where the source distribution ({\color{red}red}) is mapped to the target distribution ({\color{green}green}), with blue lines indicating the transport paths between each pair. The right diagram depicts the proposed DOT approach but only considers the marginal distribution adaptation, where both the source and target distributions are mapped onto the subspaces ({\color{brown}brown}) respectively.\label{Fig:2_2_SPD}
}
\end{figure}

\subsection{Synthetic Dataset}

We synthesize a source domain dataset consisting of SPD matrices of size $2 \times 2$, drawing a Gaussian distribution on the SPD manifold proposed in a generative approach~\cite{said2017riemannian} with a median of 1 and a standard deviation of 0.4. We adopt the following transformation matrix in the generation: 
\begin{equation*}
W := 
\begin{pmatrix}
1 & 0.5 \\
0.5 & 1 \\
\end{pmatrix},
\end{equation*}
and data in the target distribution can be expressed as $WSW^\top$.

This experiment does not generate label information, so the DOT method we use only considers marginal distribution adaptation. From Fig.~\ref{Fig:2_2_SPD}, we observe that the DOT method effectively learns a mapping $\varphi$ that aligns the source and target distributions $X_S$ and $X_T$ onto the subspaces $\varphi(X_S) = \varphi(X_T)$ by minimizing the difference between their means after mapping.

\subsection{Comparative Experiments on Real-World EEG Datasets}

As summarized in Table~\ref{tab:comparative_experiments}, we categorize all evaluated methods into three groups based on their reliance on Riemannian metrics: (1) non-metric approaches, (2) methods using the Affine-Invariant Riemannian Metric, and (3) those employing the Log-Euclidean Metric. Results demonstrate significant metric-dependent performance variations. For instance, on the BNCI2014001 dataset, EMD+SVM achieves 50.89\% accuracy under $g^{AIRM}$ versus 68.33\% under $g^{LEM}$. 

Key empirical insights emerge: (i) Participant scale: Transfer methods yield more pronounced improvements (often >5\%) in smaller cohorts due to heightened inter-subject variability; (ii) Baseline dependency: Methods achieve larger gains when base classifiers (e.g., SVM, MDRM, SPDNet) underperform initially; (iii) Methodological sensitivity: Performance correlates with architectural choices (e.g., neural network design), hyperparameter tuning, and preprocessing pipelines, limiting universal generalizability.

\begin{table*}[!h]
\caption{Experimental Results: All results in the table are expressed in percentages (\%) and include comparisons across four categories, nine methods, and their corresponding three base classifiers. The baseline results of these classifiers (base classifier), prior to the application of any transfer methods, are presented at the bottom of the table. The abbreviation "Acc." stands for average accuracy, and the symbol $\uparrow$ highlights the performance improvements achieved by incorporating transfer methods compared to the baseline results without them. 
}
\centering 

\resizebox{\textwidth}{!}{ 

\begin{tabular}{c|c|c|c|c|c|c|c|c|c}
\toprule
 \multirow{5}{*}{Metric} &
 \multirow{5}{*}{Category}& 
  \multirow{5}{*}{Method} &
  \multirow{5}{*}{Base} &  \multicolumn{2}{c|}{KU}  & \multicolumn{2}{c|}{BNCI2014001} & \multicolumn{2}{c}{BNCI2015001} \\
   &&&&  \multicolumn{2}{c|}{(54 Subjects)}  & \multicolumn{2}{c|}{(9 Subjects)} & \multicolumn{2}{c}{ (12 Subjects)} \\
& &  &&  \multicolumn{2}{c|}{$S1 \longrightarrow S2$}  & \multicolumn{2}{c|}{$T \longrightarrow E$} & \multicolumn{2}{c}{$S1 \longrightarrow S2$} \\
& &  & Classifier& \multicolumn{2}{c|}{Semi-supervised}  & \multicolumn{2}{c|}{Unsupervised} & \multicolumn{2}{c}{Unsupervised} \\
\cline{5-10}\\[-10pt]
&&&& Acc. & $\uparrow$ & Acc. & $\uparrow$  & Acc. & $\uparrow$ \\
\midrule
\multirow{6}{*}{$g^{AIRM}$} & OT  & EMD & SVM & 65.24 (14.81)& $-$0.07 & 50.89 (16.54) & $-$14.27 & 82.17 (11.80) & $+$6.25\\
\cline{2-10}\\[-10pt]
& \multirow{4}{*}{PT}  & RCT &MDRM & 52.69 (6.39) & $+$0.34 &  \multirow{2}{*}{55.13 (12.48)} & \multirow{2}{*}{$+$0.89} & \multirow{2}{*}{77.67 (13.23)} & \multirow{2}{*}{ $+$5.46}\\
  & {}  & ROT &MDRM & 51.39 (5.89) & $-$0.96 & & & & \\
  \cline{3-10}\\[-10pt]
 & {}  & RCT & SPDNet & 69.91 (13.67) & $+$2.15& \multirow{2}{*}{72.11 (12.99)} & \multirow{2}{*}{$+$3.51} & \multirow{2}{*}{81.25 (10.64)} & \multirow{2}{*}{$+$4.75}\\
 & {}  & ROT & SPDNet & 63.44 (14.94) & $-$4.32 & &&& \\
 \cline{2-10}\\[-10pt]
 & DPT  & RieBN &SPDNet & 67.57 (15.27) & $-$0.19 & 69.33 (13.49) & $+$0.73 & 78.83 (14.56) & $+$2.33\\
\midrule
\multirow{8}{*}{$g^{LEM}$}& OT & EMD & SVM & 67.70 (15.14) & $-$0.06  & 68.44 (15.79)& $+$3.28 & 82.17 (11.80)& $+$6.25\\
&& EMD & SPDNet & 64.22 (14.30) & $-$3.54 & 62.15 (14.69) & $-$6.45 &78.33 (13.80) & $+$1.83\\
&&SPDSW & SVM & 66.67 (14.54) & $+$1.36 & 67.82 (16.71) & $+$2.66 & 82.42 (11.74) & $+$6.50\\
&&logSW & SVM & 66.26 (15.00)  & $+$0.95 & 65.82 (16.56)& $+$0.66 & 80.75 (12.94) & $+$4.83\\
\cline{2-10}\\[-10pt]
&\multirow{4}{*}{DOT} &DeepJDOT & SPDNet & 67.38 (14.67) & $-$0.38 & 61.00 (11.40) & $-$7.60 & 77.63 (15.12) & $+$1.13\\
&        &  MDA &SPDNet & 69.19 (14.37) & $+$1.43 & 67.78 (12.46) & $-$0.82 & 78.33 (14.57)& $+$1.83\\
&	  & CDA & SPDNet & 68.80 (14.29) & $+$1.04 & 65.12 (11.34) & $-$3.48 & 78.25 (14.59) & $+$1.75\\
&	  & MDA/CDA & SPDNet & 68.61 (14.79)& $+$1.05 & 64.85 (11.54) & $-$3.75 & 78.29 (11.99) & $+$1.79\\
\midrule 
\multicolumn{3}{c|}{} & SVM & 65.31 (14.42) & & 65.16 (17.00)& & 75.92 (16.29)&\\
\multicolumn{3}{c|}{Baseline Classifer} & MDRM & 52.35 (6.73)& & 54.24 (13.31) && 72.21 (14.67)&\\
\multicolumn{3}{c|}{} &SPDNet& 67.76 (14.55)  && 68.60 (13.03) && 76.50 (14.41)&\\
\bottomrule
\end{tabular}

}

\label{tab:comparative_experiments}
\end{table*}

For methods based on the affine-invariant Riemannian metric, the recentering technique (RCT) consistently achieves strong performance across datasets, especially when combined with MDRM or SPDNet classifiers. In contrast, EMD exhibits high variability across datasets, indicating limited robustness. Within the RPA framework, the rotation step (ROT) underperforms its recentering counterpart and is further constrained by its semi-supervised nature. RieBN yields only marginal improvements and fails to consistently outperform RCT.

Under the Log-Euclidean metric, the performance of OT-based methods diverges notably. EMD combined with SVM surpasses its combination with SPDNet, while SPDSW and logSW outperform the base SVM by 1$\sim$2\%, with SPDSW consistently achieving the best results. MDA shows slight advantages over CDA on two datasets (1$\sim$2\% improvement), possibly due to CDA’s reliance on unstable pseudo-labels generated by Graph-CSPNet~\cite{ju2023graph}. In particular, DeepJDOT performs poorly on our proposed method. Although our approach simplifies alignment by minimizing transport costs between frequency-band centroids, trading off direct accuracy for computational efficiency, it significantly improves scalability without compromising robustness.

Overall, neural network-based approaches consistently outperform traditional methods such as EMD and SPDSW. Although feature engineering remains critical for SVM-based techniques, deep learning models benefit from the representations learned. Although public implementations of SPDSW and logSW yield better results than ours on BNCI2014001 - likely due to task-specific preprocessing - their dependence on handcrafted features highlights a key limitation compared to the flexibility and generalizability of deep models.

\section{DISCUSSIONS}~\label{sec:discussion}

In the following discussion, we address several critical limitations inherent to our proposed methodology:

\subsection{Theorem~\ref{t1} with cost function $c(x,\bar{x})=||x-\bar{x}||_{\ell^2}^2$}~\label{appen:euclidean_cost}
We claim that Theorem~\ref{t1} with cost function $c(x,\bar{x})=||x-\bar{x}||_{\ell^2}^2$ is equilvant to~\cite[Theorem 3.1]{courty2016optimal}.
It was first observed and proved by Yair et al.\cite{yair2019domain} who use the matrix vectorization operator to establish an isometry between the affine transformation and the BiMap transformation. 
We follow their arguments and revise a bit of their proof. 
\begin{theorem}\cite[Theorem 3.1]{courty2016optimal}~\label{t2}
Let $\mu$ and $\nu$ be two discrete distributions on $\mathbb{R}^n$ with $N$ Diracs.
Given a strictly positive definite matrix $A$, bias weight $b\in\mathbb{R}^n$, and source samples $\{x_i\}_{i=1}^N \sim \mu$, suppose target samples $\bar{x}_i:= A x_i + b$, for $i=1, ..., N$, and the weights in both source and target distributions' empirical distributions are $1/N$. 
Then, transport $T(x_i) = A x_i + b$ is the solution to the OT Problem provided with a cost function $c(x, \bar{x}):= ||x - \bar{x}||_{\ell^2}^2$.
\end{theorem}
To expose the relationship between the BiMap transformation in Theorem~\ref{t1} and affine transformation in Theorem~\ref{t2}, we introduce the matrix vectorization operator $vec(\cdot)$ to stack the column vectors of matrix $A=(a_1 \big| a_2 \big| \cdots \big| a_n)$ below on another as follows, 
\begin{equation*}
vec(A) := 
\begin{pmatrix}
a_1 \\
a_2 \\
\vdots \\
a_n \\
\end{pmatrix}.
\end{equation*}
Then, we have the following lemma.
\begin{lemma}~\label{lemma_vec_1}
$vec(WSW^{\top}) = (W \otimes W) vec(S)$, where $\otimes$ is Kronecker product.
\end{lemma}

\begin{proof}
Suppose $S$ is an $n\times n$ matrix, and $W^\top= (w_1^\top| w_2^\top|\cdots | w_m^\top)$ is the transpose of $m\times n$ matrix $W$ with each column vector $w_i^\top \in \mathbb{R}^n$ (i=1, ..., m).
The k-th column vector of $WSW^{\top}$ can be written as $(WSW^{\top})_{(:, k)} = WSw_k^\top = (w_k^\top \otimes W)vec(S)$.
Hence, we achieve the lemma as follows, 
\begin{align*}
vec(WSW^{\top}) 
&= 
\begin{pmatrix}
(WSW^{\top})_{(:, 1)} \\
\vdots \\
(WSW^{\top})_{(:, m)} \\
\end{pmatrix} \\
&=
\begin{pmatrix}
w_1^\top \otimes W\\
\vdots \\
w_m^\top \otimes W\\
\end{pmatrix} vec(S) \\
&= (W \otimes W) vec(S).\\
\end{align*}
\end{proof}

\begin{lemma}~\label{lemma_vec_2}
Suppose $W \in \mathbb{R}^{n\times n}$ is positive definite, then $W \otimes W$ is also positive definite.  
\end{lemma}

\begin{proof}
This is given by a fact that suppose $\{\lambda_1, ..., \lambda_n\}$ are the eigenvalues of $W$, then the eigenvalues for $W \otimes W$ is $\{\lambda_1^2, ..., \lambda_n^2\}$. 
Hence, $W \otimes W$ is positive definite. 
\end{proof}

According to Lemma~\ref{lemma_vec_1} and~\ref{lemma_vec_2}, the BiMap transformation $W\cdot S \cdot W^\top$ turns to be affine transformation $A = W \otimes W$, and thus Theorem~\ref{t1}. 
Hence, Theorem~\ref{t1} endowed with the cost function $c\big(vec(x), vec(y)\big) = ||vec(x) - vec(y)||_{\ell^2}^2$ is equilvant to Theorem~\ref{t2}.

\subsection{Parallel Transport Under Log-Euclidean Metric}

Parallel transport on Riemannian manifolds transfers a tangent vector at one point along a curve on the manifold to another point, such that the transported vector remains parallel to the original vector~\cite{petersen2006riemannian}. 
Under two Riemannian metrics on SPD manifolds $g^{AIRM}$ and $g^{LEM}$, expressions of parallel transport exhibit distinctly and determine two distinct modeling approaches for the problems~\cite{sra2015conic,yair2019domain}. 
Specifically, for any $S_1$ and $S_2$ on ($\mathcal{S}_{++}$, $g^{AIRM}$), parallel transport from $S_1$ to $S_2$ for $s \in T_{S_1} \mathcal{S}_{++}$ can be expressed as $\Gamma_{S_1 \rightarrow S_2} (s) = E s E^\top$, where $E = (S_1 \cdot S_2^{-1})^{\frac{1}{2}}$. 
For any $S_1$ and $S_2$ on ($\mathcal{S}_{++}$, $g^{LEM}$), parallel transport is given by $\Gamma_{S_1 \rightarrow S_2} (s) = s$. 

The DOT framework seeks a neural networks-based transformation $\varphi: X_S, X_T \longmapsto Z \subset \mathcal{S}_{++}$ such that $\mathbb{P}_S (\varphi(X_S)) = \mathbb{P}_T(\varphi(X_T))$ and $\mathbb{Q}_S(Y_S|\varphi(X_S)) = \mathbb{Q}_T(Y_T|\varphi(X_T))$, where $\varphi(X_S)$ and $\varphi(X_T)$ are transformation $\varphi$ on the source and target feature spaces. 
Neural networks-based transformation $\varphi$ extracts discriminative information from feature spaces. 
After the extraction, it is reasonable to view the feature space of the source domain as the same as it of the target domain, as they will be transformed under neural networks with the same weights, so are their transformed feature spaces, i.e., $X_T = X_S$ and $\varphi(X_T) = \varphi(X_S) \subset Z$, and thus we utilize the Log-Euclidean metric to formulate the space of covariance matrice.

\subsection{Pseudo-Label-Driven Conditional Distribution Adaptation}

In this study, we adopt a pseudo-labeling strategy inspired by the JDA framework~\cite{long2013transfer} to facilitate the alignment of conditional distributions between source and target domains. Specifically, we assign pseudo-labels to the unlabeled target domain samples using a base classifier trained on labeled source domain data. While discrepancies in both marginal and conditional distributions may introduce labeling noise, this strategy provides a practical foundation for conditional alignment in the absence of target labels. For EEG classification tasks, we employ Graph-CSPNet as the base classifier, as it outperforms the best in this setting~\cite{ju2023graph}. Importantly, this pseudo-labeling mechanism not only enables conditional distribution alignment but also improves cross-domain adaptation by capturing domain-specific discriminative features.

\subsection{Time Complexity in Optimal Transport}

The computational complexity of the discrete OT problem poses significant challenges, especially for high-dimensional or large-scale datasets. Traditional methods for the discrete OT problem—such as the simplex algorithm or the Hungarian method—often require $O(N^3)$ time when solving an $N\times N$ assignment problem, limiting their application to small-scale scenarios~\cite{peyre2019computational}. Entropy-regularized approaches, such as the Sinkhorn algorithm, reduce the computational burden by achieving a per-iteration complexity of $O(KNM)$ (which becomes $O(KN^2)$ when $N = M$) over $K$ iterations~\cite{cuturi2013sinkhorn}. For high-dimensional settings, techniques like sliced Wasserstein distances—which project data onto $L$ random one-dimensional subspaces—have a computational complexity of $O(LN\log N)$~\cite{bonnotte2013unidimensional}.

In practical applications, trade-offs are essential. While exact methods are suitable for small, controlled problems, in EEG classification, the decision to use Wasserstein barycenters under the Log-Euclidean metric as intermediate reference distributions with neural networks is guided by specific application requirements. The Barycenter under the Log-Euclidean metric, which has been simplified as the arithmetic mean of their tangent representation, proves particularly beneficial in multi-source domain adaptation scenarios (for example, cross-session BCI calibration) by reducing computational overhead and enhancing stability.

\subsection{Average Log-Euclidean Distance in Assumption 3}

This section provides empirical validation for the reasonableness of Assumption~\ref{a3} through numerical analysis. Table~\ref{tab:LEM_distance} presents the average Log-Euclidean distances between frequency bands across domains. Notably, the diagonal entries—representing distances between the same frequency bands in the source and target domains—consistently exhibit the smallest values within each row for both datasets. This observation validates that our mapping successfully aligns the center of each source domain frequency band with its corresponding target domain counterpart. The results align with the intuitive expectation that covariance structures within the same frequency band (e.g., alpha or beta bands) should show smaller divergences due to shared neurophysiological characteristics, whereas comparisons across different frequency bands naturally result in larger distances.

The computation protocol is as follows: Given source and target domain samples structured as [trials × frequency bands × channels × channels], we first compute the Fréchet mean for each frequency band by averaging the covariance matrices across trials, resulting in a [frequency bands × channels × channels] representation. We then compute pairwise Log-Euclidean distances between all frequency band pairs, where intra-band distances (on the diagonal) reflect the accuracy of alignment, while inter-band distances (off-diagonal) capture the divergence between different frequency bands.

\begin{table*}
\caption{Average Log-Euclidean distances between the Fréchet means of EEG spatial covariance matrices generated from different frequency bands in the source and target domains for the KU (54 subjects), BNCI2014001 (9 subjects), and BNCI2015001 (12 subjects) datasets. 
The columns and rows are the sequence of frequency bands. 
The shortest average Log-Euclidean distance in each row is highlighted in boldface.
}
\centering 

\resizebox{0.9\textwidth}{!}{ 

\begin{tabular}{rrrrrrrrrr}
\toprule
KU &&&&&&&&&\\
S1 $\backslash$ S2   &  4$\sim$8  Hz&     8$\sim$12  Hz&    12$\sim$16 Hz&    16$\sim$20 Hz&    20$\sim$24 Hz&    24$\sim$28 Hz&    28$\sim$32 Hz&     32$\sim$36 Hz&  36$\sim$40 Hz\\
\midrule
 4$\sim$8 Hz& \textbf{4.69} & 4.98 & 5.82 & 6.74 & 7.19 & 7.64 & 8.17 & 8.78 & 9.28 \\
8$\sim$12 Hz & 5.09 & \textbf{4.21} & 4.82 & 5.81 & 6.25 & 6.70 & 7.22 & 7.82 & 8.31 \\
12$\sim$16 Hz & 5.84 & 4.76 & \textbf{4.01} & 4.62 & 4.98 & 5.34 & 5.78 & 6.30 & 6.75 \\
16$\sim$20 Hz & 6.66 & 5.61 & 4.37 & \textbf{3.96} & 4.08 & 4.33 & 4.64 & 5.02 & 5.40 \\
 20$\sim$24 Hz & 7.08 & 6.01 & 4.66 & 3.97 & \textbf{3.86} & 3.99 & 4.27 & 4.63 & 4.97 \\
 24$\sim$28 Hz & 7.52 & 6.45 & 4.98 & 4.11 & 3.86 & \textbf{3.75} & 3.91 & 4.21 & 4.52 \\
 28$\sim$32 Hz & 8.04 & 6.96 & 5.40 & 4.37 & 4.07 & 3.82 & \textbf{3.73} & 3.87 & 4.11 \\
32$\sim$36 Hz  & 8.65 & 7.56 & 5.93 & 4.77 & 4.42 & 4.13 & 3.88 & \textbf{3.74} & 3.82 \\
36$\sim$40 Hz & 9.16 & 8.07 & 6.40 & 5.16 & 4.77 & 4.43 & 4.11 & 3.82 & \textbf{3.74} \\
\bottomrule
\toprule
BNCI2014001&&&&&&&&&\\
T $\backslash$ E   &  4$\sim$8 Hz&     8$\sim$12 Hz&    12$\sim$16 Hz&    16$\sim$20 Hz&    20$\sim$24 Hz&    24$\sim$28 Hz&    28$\sim$32 Hz&  32$\sim$36 Hz& 36$\sim$40 Hz\\
\midrule
 4$\sim$8 Hz   & \textbf{2.81} &  4.25 & 5.14 & 5.88 & 6.70 & 8.30 & 9.64 & 10.97 & 12.46 \\
8$\sim$12 Hz  & 4.31 &  \textbf{2.37} & 3.98 & 5.45 & 6.28 & 8.07 & 9.44 & 10.77 & 12.26 \\
12$\sim$16 Hz &5.16 &  3.91 & \textbf{2.29} & 3.53 & 4.23 & 5.83 & 7.15 &  8.42 &  9.87 \\
16$\sim$20 Hz  & 5.84 &  5.49 & 3.44 & \textbf{2.17} & 2.76 & 4.08 & 5.29 &  6.56 &  7.98 \\
 20$\sim$24 Hz  & 6.59 &  6.24 & 4.18 & 2.75 & \textbf{2.18} & 3.31 & 4.54 &  5.79 &  7.24 \\
 24$\sim$28 Hz  &8.11 &  8.02 & 5.72 & 4.01 & 3.14 & \textbf{2.20} & 2.96 &  4.03 &  5.38 \\
 28$\sim$32 Hz  &9.42 &  9.38 & 7.06 & 5.21 & 4.39 & 2.83 & \textbf{2.21} &  2.86 &  4.05 \\
32$\sim$36 Hz   &10.67 & 10.63 & 8.25 & 6.43 & 5.57 & 3.87 & 2.72 &  \textbf{2.22} &  3.05 \\
36$\sim$40 Hz   &12.16 & 12.12 & 9.69 & 7.85 & 7.02 & 5.18 & 3.85 &  2.70 &  \textbf{2.22} \\
\bottomrule
\toprule
BNCI2015001&&&&&&&&&\\
 $S_1$ $\backslash$ $S_2$  &  4$\sim$8 Hz&  8$\sim$12 Hz&  12$\sim$16 Hz& 16$\sim$20 Hz& 20$\sim$24 Hz& 24$\sim$28 Hz& 28$\sim$32 Hz&  32$\sim$36 Hz& 36$\sim$40 Hz\\
\midrule
  4$\sim$8 Hz  & \textbf{1.07} & 12.39 & 12.35 & 10.36 &  9.71 &  8.46 &  5.73 &  2.25 &  6.24 \\
8$\sim$12 Hz  & 12.04 &  \textbf{1.06} &  1.49 &  2.68 &  3.22 &  4.40 &  7.13 & 12.43 & 18.19 \\
12$\sim$16 Hz & 12.10 &  1.60 &  \textbf{1.09} &  2.52 &  3.13 &  4.32 &  7.08 & 12.41 & 18.18 \\
16$\sim$20 Hz & 10.19 &  2.99 &  2.67 &  \textbf{1.19} &  1.59 &  2.45 &  5.06 & 10.40 & 16.22 \\
20$\sim$24 Hz & 9.64 &  3.40 &  3.15 &  1.53 &  \textbf{1.22} &  1.90 &  4.51 &  9.84 & 15.66 \\
24$\sim$28 Hz &  8.36 &  4.51 &  4.30 &  2.34 &  1.80 &  \textbf{1.19} &  3.21 &  8.52 & 14.34 \\
 28$\sim$32 Hz  & 5.72 &  7.17 &  7.01 &  4.94 &  4.31 &  3.06 &  \textbf{1.28} &  5.73 & 11.54 \\
32$\sim$36 Hz & 2.52 & 12.43 & 12.31 & 10.24 &  9.58 &  8.29 &  5.44 &  \textbf{1.30} &  6.17 \\
36$\sim$40 Hz  & 6.41 & 18.22 & 18.12 & 16.09 & 15.42 & 14.14 & 11.27 &  5.88 &  \textbf{1.25} \\
\bottomrule
\end{tabular}
}
\label{tab:LEM_distance}
\end{table*}

\section{CONCLUSIONS}

This study extends the OT-DA framework to operate on SPD manifolds equipped with the Log-Euclidean metric. We propose a novel geometric deep learning approach for domain adaptation with SPD matrix-valued data by aligning both marginal and conditional distributions. The method specifically addresses variability between sessions in EEG-based motor imagery classification - a long-standing challenge in brain-computer interface research, where changes in signal distributions between sessions often degrade model performance. Experiments on three public EEG datasets (KU, BNCI2014001, BNCI2015001) validate the effectiveness of the proposed method, showing that it can effectively reduce distribution shifts between sessions. 

Despite these contributions, the approach still faces several limitations. One significant challenge is the sensitivity to errors in pseudo-labels, even though the alignment of conditional distributions is stabilized using sufficient statistics. Early-stage label predictions, which often have around 70\% accuracy in cross-session EEG settings, can hinder adaptation due to error propagation. This issue limits the overall effectiveness of the adaptation process. Furthermore, while transfer learning methods show measurable effectiveness in improving classification performance, domain change is not the only determinant of performance. As a result, the improvements from transfer techniques are often modest. A major limitation is the lack of a robust quantitative framework to measure the intensity of domain shift and its systematic relationship with classifier performance. This gap prevents a comprehensive evaluation of the true efficacy of the proposed methodology. Future work should focus on addressing these challenges, particularly by enhancing the robustness of pseudo-labeling through uncertainty-aware alignment strategies.

\section*{Acknowledgment}
This work was supported under the RIE2020 Industry Alignment Fund–Industry Collaboration Projects (IAF-ICP) Funding Initiative, as well as cash and in-kind contributions from industry partner(s); This work was also supported by the RIE2020 AME Programmatic Fund, Singapore (No. A20G8b0102).





\section*{Declaration of generative AI and AI-assisted technologies in the writing process}

During the preparation of this work, the author(s) used ChatGPT 4 (OpenAI) in order to improve the English grammar of this manuscript. After using this tool/service, the author(s) reviewed and edited the content as needed and take(s) full responsibility for the content of the publication.

\bibliographystyle{elsarticle-num}
\bibliography{refs}







\end{document}